\documentclass[letter, 12pt, onecolumn]{ieeetran}
\IEEEoverridecommandlockouts
\usepackage{graphicx}
\usepackage{eqlist}
\usepackage{amsfonts}
\usepackage{tabularx}
\usepackage{booktabs}
\usepackage{amssymb}
\usepackage{amsmath}
\let\labelindent\relax
\usepackage{enumitem}
\usepackage{threeparttable}
\usepackage[lined, ruled, linesnumbered, commentsnumbered]{algorithm2e}
\usepackage{lipsum}

\usepackage[usenames,dvipsnames]{xcolor}
\usepackage{comment}
\usepackage{flushend}
\usepackage{todonotes}
\usepackage{xcolor}
\usepackage{booktabs}

\begin{document}
\title{Implicit Contact-Rich Manipulation Planning for\\a Manipulator with Insufficient Payload}

\author{Kento Nakatsuru$^{1}$, Weiwei Wan$^{\ast1}$, and Kensuke Harada$^{12}$
\thanks{$^\ast$ Weiwei Wan is the corresponding author. Email: wan@sys.es.osaka-u.ac.jp. $^1$ Graduate School of Engineering Science, Osaka University, Osaka, Japan. $^2$ National Institute of Advanced Industrial Science and Technology (AIST), Tsukuba, Japan.}}
\maketitle

\begin{abstract}
\noindent\textbf{Purpose of this paper:}
This paper studies using a mobile manipulator with a collaborative robotic arm component to manipulate objects beyond the robot's maximum payload.\\
\textbf{Design/methodology/approach:}
The paper proposes a single-short probabilistic roadmap-based method to plan and optimize manipulation motion with environment support. The method uses an expanded object mesh model to examine contact and randomly explores object motion while keeping contact and securing affordable grasping force. It generates robotic motion trajectories after obtaining object motion using an optimization-based algorithm. With the proposed method’s help, we can plan contact-rich manipulation without particularly analyzing an object’s contact modes and their transitions. The planner and optimizer determine them automatically.\\
\textbf{Findings:}
We conducted experiments and analyses using simulations and real-world executions to examine the method's performance. The method successfully found manipulation motion that met contact, force, and kinematic constraints. It allowed a mobile manipulator to move heavy objects while leveraging supporting forces from environmental obstacles.\\
\textbf{What~is~original/value~of~paper:}
The paper presents an automatic approach for solving contact-rich heavy object manipulation problems. Unlike previous methods, the new approach does not need to explicitly analyze contact states and build contact transition graphs, thus providing a new view for robotic grasp-less manipulation, non-prehensile manipulation, manipulation with contact, etc.
\end{abstract}

\section{Introduction}
\begin{figure}[!htpb]
  \begin{center}
  \includegraphics[width=\linewidth]{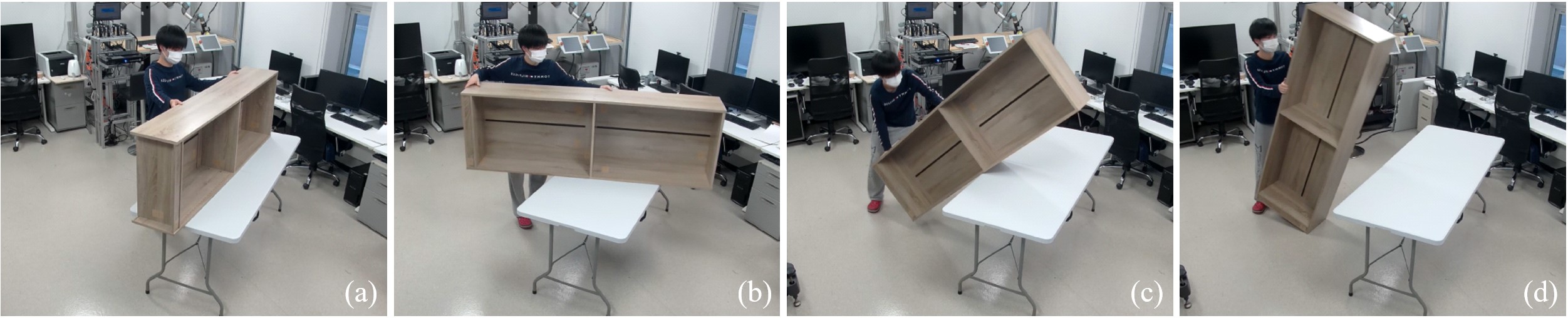}
  \caption{A human moved a large and heavy shelf down from a table. The human took advantage of environmental support to (a, b) push, (c) slide, and (d) pivot the shelf. It is a sequence of contact-rich manipulation actions. This paper aims to develop a generalized planning and optimization method for a mobile manipulator to handle objects like the human.}
  \label{fig:ts}
  \end{center}
\end{figure}

This paper studies using a mobile manipulator to manipulate objects beyond the robot's maximum payload. Fig. \ref{fig:ts} exemplifies how a human manipulates a large, heavy object beyond the person's maximally affordable weight. The human took advantage of environmental support to share partial object weight and thus unloaded the object from the table onto the ground. This paper develops a planning and optimization approach for a mobile manipulator to handle objects like the human in the figure. The approach uses an expanded object mesh model to examine contact and randomly explore object motion while keeping contact and securing affordable grasping force. It generates robotic motion trajectories after obtaining object motion using an optimization-based algorithm. With the proposed methods' help, we can plan contact-rich manipulation without particularly analyzing an object's contact modes and their transitions, and thus allow a mobile manipulator to move heavy objects while leveraging supporting forces from environmental obstacles.

Handling large and heavy objects are challenging for robotic manipulators, as high-quality motor and reducer modules and careful mechanical design are needed to increase a robotic manipulator's payload. The module and design costs make heavy-duty robots unaffordable for small and medium manufacturers. On the other hand, collaborative manipulators are safe and can be quickly and flexibly deployed to manufacturing sites. They have a limited payload and cannot lift and move heavy objects that exceed the limitation. Previously, researchers have been developing methods for collaborative manipulators to handle large and heavy objects. The methods took advantage of environmental support to share partial object weight. The robot manipulator performs non-prehensile actions to push, flip, or pivot large and heavy objects \cite{fakhari2021}\cite{zhang2021}\cite{hayakawa2021dual}\cite{fan19}. Using collaborative manipulators while considering environmental support allowed the manipulation of large and heavy objects with affordable hardware costs.

However, the previous methods used uniform sampling and search graphs to model the object-environment and object-robot contact states. The methods inherently have completeness problems. Whether a contact and grasping state will be considered for exploration and optimization depends on if the state was encoded in the graph. The methods cannot exploit an unmodeled state to achieve given goals. 

Unlike previous methods, we propose a single-short probabilistic contact-rich planning method to explore contact states that meet force constraints automatically. The method is based on the Rapidly-exploring Random Tree (RRT) routine. It randomly explores object motion while maintaining object-environment contact. The contact is not defined as a limited number of points, edges, or faces. It is instead maintained using an expanded contact mesh and may change freely as long as the ``crust'' formed by the expanded object mesh is in touch with the environment. The expanded mesh is determined functionally by considering the mechanical compliance of a manipulator. Robotic motion are generated using an optimization-based algorithm based on the probabilistically planned object motion. With the proposed method's help, we can find contact-rich manipulation without explicitly analyzing an object's contact states and their transitions. The planner and optimizer determine them automatically.

We conducted experiments using a mobile manipulator to analyze the proposed method's performance. The method could help find contact-rich manipulation trajectories allowing the mobile manipulator to move large, heavy objects with environmental support. The optimizer ensured the manipulator's joints bore affordable forces not to damage themselves. The proposed method is expected to be an important contribution to contact-rich manipulation and starts a new view for solving related problems.

The organization of this paper is as follows. Section II reviews related work. Section III presents an overview of the proposed method using a schematic diagram. Section IV$\sim$V show algorithmic details like the expanded collision models, RRT-based planning, and robotic trajectory optimization. Section V presents experiments and analysis. Conclusions and future work are drawn and discussed in Section VI.

\section{Related Work}
\label{sec-related}
\subsection{Manipulating Heavy Objects}
Using robots to manipulate heavy objects is a challenging topic \cite{ohashi2016realization}. In contemporary studies, people developed special-purpose mechatronic systems, manipulators with a large payload \cite{du2022}, robot-human collaboration \cite{kim18}\cite{stouraitis2020online}, robot-machine collaboration \cite{hayakawa2021adr}\cite{recker21}\cite{ikeda18}\cite{balatti20}, non-prehensile manipulation policies \cite{yoshida2010pivoting}\cite{specian18}\cite{nazir21}, etc., to solve the problems.

Particularly, O'Neill et al. \cite{oneill21} developed an autonomous method for safely tumbling a heavy block sitting on a surface using a two-cable crane. Kayhani et al. \cite{kayhani21} developed an automated lift path planning method for heavy crawler cranes in no-walk scenarios employing a robotics approach. Du et al. \cite{du2022} developed a high-duty manipulator to replace heavy disc cutters installed on the head of a tunnel boring machine. Kim et al. \cite{kim18} proposed a novel human–robot collaboration (HRC) control approach to alert and reduce a human partner's static joint torque overloading while executing shared tasks with a robot.

In this work, we employ environmental contact as essential support \cite{patankar2020} to manipulate heavy objects. In order to realize the manipulation of objects that exceed the payload, we optimized the robot motion by considering contacted object trajectories, supporting forces, manipulator poses, and the load on each joint. One advantage of using the environment is that no additional equipment or facilities are required. While additional equipment can be very effective in accomplishing specific tasks, issues such as loss of versatility still need to be addressed. For example, using a cart to load and move objects makes it very difficult to maneuver in an environment with steps \cite{ohashi16}. Also, when robots are integrated with additional equipment, it is necessary to solve a highly constrained closed-chain problem that occurs when robots work together, which results in a considerable restriction on the IK range and requires planning for re-grasping. In contrast, taking advantage of support from environmental obstacles provides more freedom for maintaining contact. With the proposed object trajectory planning and robot motion optimization methods in this work, a robot can find a contact-rich manipulation motion without analyzing an object's contact modes and transitions and thus move heavy objects while leveraging supporting forces from environmental obstacles.

\subsection{Manipulation Considering Changing Contact Modes}

Contact modes have been an old topic in robotic manipulation \cite{hirukawa94}\cite{yu96}\cite{ji01}\cite{aiyama01}\cite{maeda2005planning}. Modern studies used contact modes and mode transition graphs to guide probabilistic planning or optimization. For example, Raessa et al. \cite{raessa2021planning} developed a hierarchical motion planner for planning the manipulation motion to repose long and heavy objects considering external support surfaces. Cheng et al. \cite{cheng2022contact} proposed Contact Mode Guided Manipulation Planning (CMGMP) for 3D quasistatic and quasi-dynamic rigid body motion planning in dexterous manipulation. Murooka et al. \cite{murooka2017global} proposed a planning method of whole-body manipulation for various manipulation strategies. Hou et al. \cite{hou2020} proposed a method to select the best-shared grasp and robot actions for a desired object motion. Some others do not perform detailed planning and leave contact mode selection to optimization. For instance, Sleiman et al. \cite{sleiman2019contact} presented a reformulation of a Contact-Implicit Optimization (CIO) approach that computes optimal trajectories for rigid-body systems in contact-rich settings. Aceituno-Cabezas et al. \cite{aceituno2020global} proposed a global optimization model (CTO, Contact-Trajectory Optimization) for this problem with alternated-sticking contact.

In this work, we develop a planner that implicitly takes advantage of environment contact and switches among contact modes to manipulate heavy objects. To enable this contact mode switching, we introduce an extended contact model. This allows objects to change freely, with contact not being fixed to a limited number of edges and faces, and enables planning of contact-rich path panning without detailed analysis of object contact modes and their transitions.

\section{Schematic View of the Proposed Method}

\begin{figure}[!htpb]
  \begin{center}
  \includegraphics[width=\linewidth]{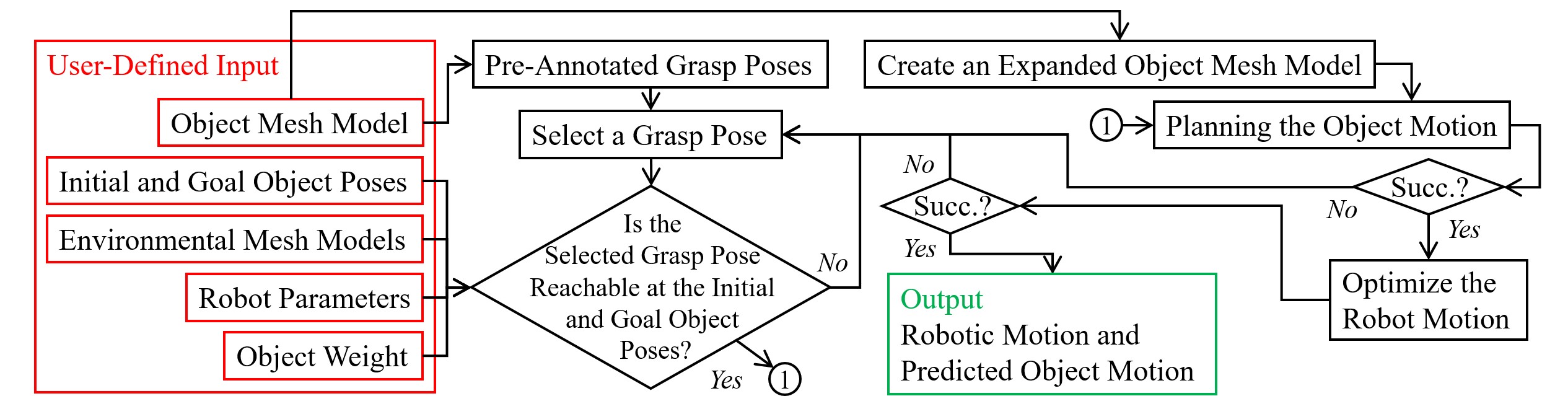}
  \caption{Schematic diagram of the proposed method.}
  \label{fig:wf}
  \end{center}
\end{figure}

Fig. \ref{fig:wf} shows the schematic diagram of the proposed method. The red box represents the user-defined input. The green box is the output. The proposed method accepts ``Object Mesh Model'', ``Initial and Goal Object Poses'', ``Environmental Mesh Models'', ``Robot Parameters'', and ``Object Weight'' as the input. It first computes a set of pre-annotated grasp poses and creates an expanded mesh using the ``Object Mesh Model'' input data. The pre-annotated grasp poses will be used for incremental planning and optimization. The method incrementally selects a grasp pose, determines if it is accessible at the initial and goal object poses, and plans the object motion and optimizes the robot trajectory. When planning the object motion, the method uses the ``crust'' formed by the expanded object mesh model and the original one to ensure continuous contact. The planner also examines the force born by the selected grasp pose at each randomly sampled roadmap node to make sure the robot can hold and move the object.

The planning part is not necessarily successful due to strict constraints. When failed, the workflow will return to the ``Select a Grasp Pose'' stage to try different grasp poses until a feasible object motion is obtained. Once a feasible object motion is obtained, the workflow will switch to the ``Optimize the Robot Trajectory'' stage to generate robotic motion trajectories. The final robotic motion trajectories may not involve the IK solutions obtained in the ``Plan the Object Motion'' stage. It is formulated as an optimization problem to minimize the robotic movement while considering the selected grasp pose and its motion in association with the object. Like the planning stage, the optimization is not necessarily successful. It will return to the ``Select a Grasp Pose'' stage in case of failure. If all stages were performed successfully, the method would produce an optimized robotic motion and the predicted object trajectory under the robotic motion as the output.


\section{Planning Object Motion}

\subsection{Expanding Models Considering Mechanical Compliance}

\begin{figure*}[!htpb]
  \begin{center}
  \includegraphics[width=\linewidth]{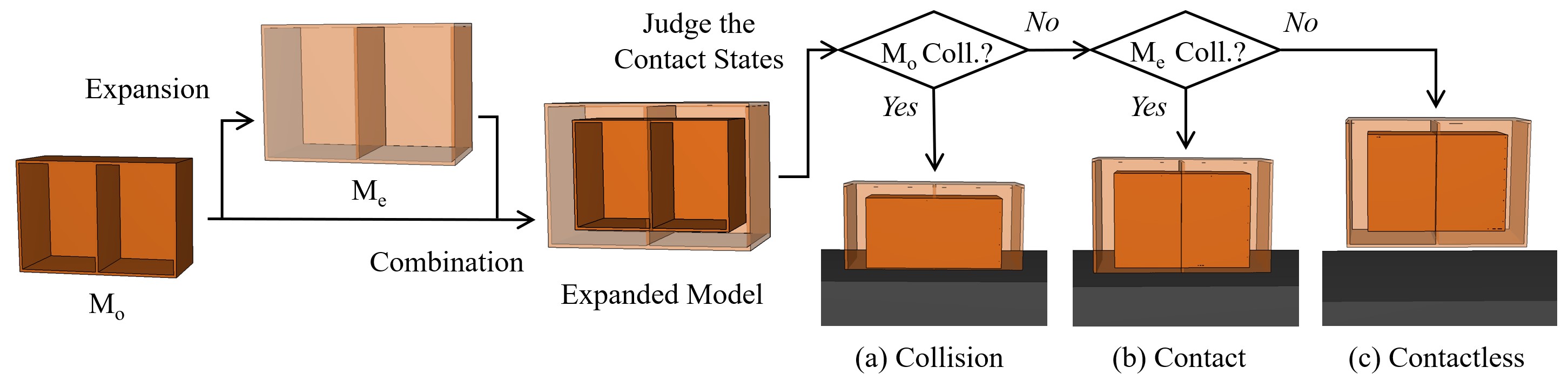}
  \caption{Concept of the expanded contact detection model.}
  \label{fig:concept}
  \end{center}
\end{figure*}

The concept behind expanding object mesh models is to build a ``crust'' and judge if the object is in contact with the environment by examining if the ``crust'' overlaps. Fig. \ref{fig:concept} illustrates the concept. Here, $\textnormal{M}_\textnormal{o}$ indicates the original object mesh model. $\textnormal{M}_\textnormal{e}$ indicates the ``crust''. We judge the relationship between the object and the environment by recognizing the overlap between $\textnormal{M}_\textnormal{o}$ and the environment and $\textnormal{M}_\textnormal{e}$ and the environment. The object is considered to be in contact with the environment when $\textnormal{M}_\textnormal{e}$ overlaps with it, while $\textnormal{M}_\textnormal{o}$ does not. If both $\textnormal{M}_\textnormal{o}$ and $\textnormal{M}_\textnormal{e}$ overlap with the environment, we consider the object is in a collision.

\begin{figure*}[!htpb]
  \begin{center}
  \includegraphics[width=\linewidth]{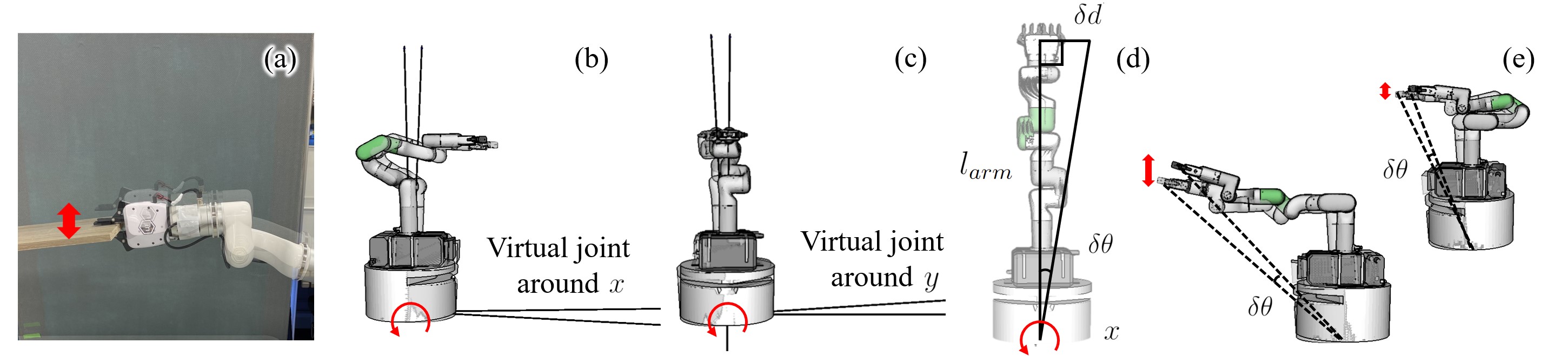}
  \caption{(a) Mechanical compliance of a mobile manipulator. (b, c) Two virtual joints. (d) Trigonometric relation about the virtual joint around the local $x$ direction. (e) Changes of compliance at different robot poses.}
  \label{fig:compliance}
  \end{center}
\end{figure*}

The thickness of the ``crust'' is crucial for the assumption to be valid. We define it by considering the mechanical compliance of a mobile manipulator. Since a mobile manipulator is not fixed and the mobile base is not entirely stiff due to springs installed at the wheels, there is slight mechanical compliance at the Tool Center Point (TCP) of the arm component. Fig. \ref{fig:compliance}(a) illustrates the compliance. We compute the compliant values by assuming two orthogonal virtual rotation joints at the base of the mobile manipulator. The two virtual rotational joints will induce absolute compliance, and the ``crust'' thickness can be computed by considering the rotation ranges of the virtual joints. Fig. \ref{fig:compliance}(b) and (c) illustrates the two orthogonal virtual rotation joints.

We set the mobile manipulator in a vertical pose, applied forces to its TCP, and measured the TCP's displacements to obtain the rotation ranges. Particularly, we computed the rotation ranges of the virtual joints using the trigonometric relations between the arm length and the displacements. Fig. \ref{fig:compliance}(d) shows the trigonometric relation about the local $x$ direction. Here, $l_{arm}$ is the distance between the assumed joint position and the TCP. $\delta d$ is the measured hand displacement. The rotation range around the local $x$ direction can be computed using 
\begin{equation}\label{eq:dt}
    \delta\theta=l_{arm}\tan^{-1}(\delta d).
\end{equation}

After obtaining the rotation ranges, we can compute the compliance of the robot hand at an arbitrary arm pose by using an $l_{arm}$ updated using forward kinematics, as shown by Fig. \ref{fig:compliance}(e). The robot will have more minor compliance when folded and larger compliance when expanded. The thickness of the ``crust'' will be determined dynamically considering the changing $\delta\theta$.


\subsection{RRT-Based Object Motion Planning}

The object motion is planned using RRT as the backbone. When sampling new nodes using the RRT backbone, we will examine the following constraints to secure continuous support forces from the environment as well as affordable ensuing robot manipulation.
\begin{itemize}
    \item Contact Constraint: The object at a newly sampled pose is in contact with environmental obstacles. The contact is assured by the collision and non-collision relationship of $\textnormal{M}_\textnormal{o}$, $\textnormal{M}_\textnormal{e}$, and the environmental obstacles' mesh models, as discussed in the previous subsection.
    \item Force Constraint: The minimum force required for the robot hand must be smaller than the mobile manipulator's maximum payload.
\end{itemize}

Especially for the force constraint, the thickness of $\textnormal{M}_\textnormal{e}$ is determined dynamically at each object pose. When a new pose is sampled during RRT planning, we will update $\textnormal{M}_\textnormal{e}$ following the robot's TCP position at the pose and determine if the object meets the contact constraint using the updated values.

\begin{figure}[!htpb]
  \begin{center}
  \includegraphics[width=\linewidth]{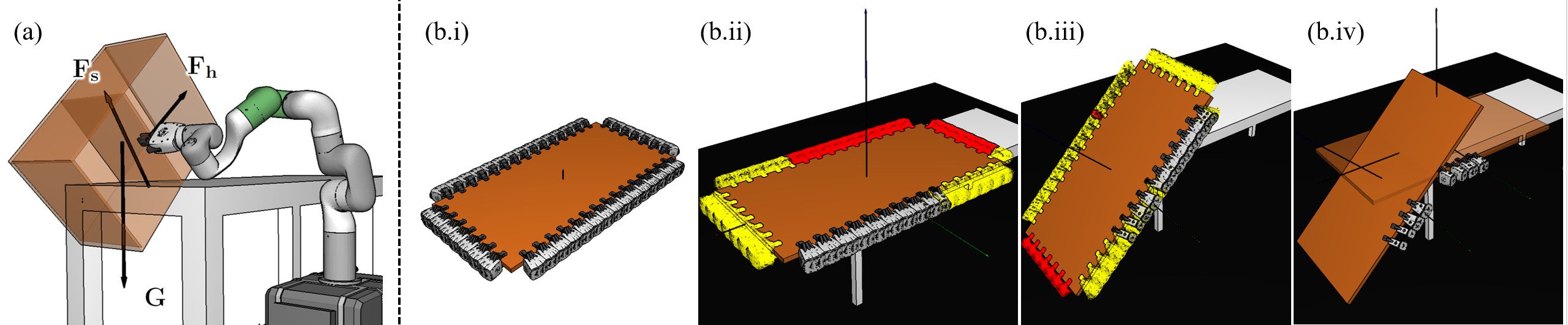}
  \caption{(a) Force constraints. (b) Determination of candidate grasping poses.}
  \label{fig:forces}
  \end{center}
\end{figure}

For a better understanding, the force constraint is illustrated in Fig. \ref{fig:forces}(a). During a contact-rich manipulation, an object is affected by gravitational force, environmental supporting forces, and robotic holding force. These forces are labeled as $\mathbf{G}$, $\mathbf{F_s}$, and $\mathbf{F_h}$ in the figure. We assume the environment is stationary and strong enough to provide enough supporting force for the target objects. There is thus no limitation on $\mathbf{F_s}$. Under this assumption, the $\mathbf{F_h}$ can be obtained by solving the following optimization problem.
\begin{subequations}\allowdisplaybreaks
    \begin{align}
        & {\small\underset{\mathbf{F_h}}{\text{min}}} & \small\mathbf{F_h}^T\mathbf{F_h} \label{eq:obj_opgoal} \\
        & {\text{s.t.}} & \small \sum\mathbf{F_s}+\mathbf{G} = 0 \label{eq:balance_sub1} \\
        && \sum\mathbf{r_s}\times\mathbf{F_s}+\mathbf{r_h}\times\mathbf{F_h}+\mathbf{r_g}\times\mathbf{G} = 0 \label{eq:balance_sub2} \\
        && \mathbf{F_s}\in\mathbf{FC_s}, ~\mathbf{F_h}^T\mathbf{F_h} \leqslant F_{max} \label{eq:cone_contact}
    \end{align}
    \label{eq:handforce}
\end{subequations}

The optimization finds the minimum force the robotic hand needs to balance the object while considering supporting forces from the contact. Equation \eqref{eq:obj_opgoal} is the optimization goal. Equation \eqref{eq:balance_sub1} and \eqref{eq:balance_sub2} are constraints for balanced forces and torques. Equation \eqref{eq:cone_contact} represents the constraints about friction cones and the maximum robot payload. A grasping pose is judged invalid if this optimization problem has no solution. The system will switch to a different candidate grasping pose and restart the object motion planning, considering the force constraints at the new grasping pose. The candidate grasping poses are planned using a grasp planner developed by the same authors \cite{wan2021tro}, as shown in Fig. \ref{fig:forces}(b.i). The manipulators' kinematic constraints are considered at the starting and goal object poses, as shown by Fig. \ref{fig:forces}(b.ii) and (b.iii). Here, the red hands indicate collided grasping poses. The yellow hands indicate the unreachable grasping poses. The hands with original colors are the feasible poses. The candidate grasping poses are the intersection of the feasible ones at the starting and goal object poses.

The result of RRT-based object motion planning is an object motion path and the associated grasp poses for each element on the path. We represent them using $\{(\mathbf{p_o}(t), \mathbf{R_o}(t))\}$ and $\{(\mathbf{p_h}(t), \mathbf{R_h}(t))\}$, where subscripts $\mathbf{o}$ and $\mathbf{h}$ represent object and hand respectively. Notation $t={1,2,...}$ in the parenthesis is a time sequence identifier. The result will be used to optimize the motion of the mobile manipulator later, as presented in the next section.


\section{Optimizing Robot Motion}

We use optimization to produce the joint-level robot motion for moving the object following the planned object motion and hand motion. The optimization problem is formulated as follows.
\begin{subequations}\allowdisplaybreaks
    \begin{align}
        & {\underset{\mathbf{q}(t)}{\text{min}}} & ||\mathbf{q}(t+1)-\mathbf{q}(t)||_2 \label{eq:rbt_opgoal} \\
        & {\text{s.t.}}
        &
        \begin{cases}
        {\mathbf{q}(t+1) = [q(i+1)^0, q(i+1)^1, ..., q(i+1)^d]^T} \\
        {\mathbf{q}(t+1)^k \in [\theta_k^-, \theta_k^+], k=JointID} \\
        \end{cases} \label{eq:joints} \\
        &&
        \begin{cases}
            \mathbf{p_r}(t+1), \mathbf{R_r}(t+1) = FK(\mathbf{q}(t+1)) \\
            ||\mathbf{p_r}(t+1)-\mathbf{p_h}(t+1)||_2\leqslant\varepsilon_p\\
            \angle(\mathbf{R_r}(t+1), \mathbf{R_g}(t+1)) \leqslant\varepsilon_R
        \end{cases} \label{eq:ik} \\
        && 
        \begin{cases}
            \boldsymbol{\tau}(t+1)=\boldsymbol{J}(\mathbf{q}(t+1))^T\mathbf{F}(t+1)\\
            \tau^k(t+1)\in[\tau_k^-, \tau_k^+], k=JointID
        \end{cases} \label{eq:joint_tq} \\
        && 
        \begin{cases}
        ||q^0(t+1)-q^0(t)||\leqslant d_x\\
        ||q^1(t+1)-q^1(t)||=0\\
        ||q^2(t+1)-q^2(t)||=0
        \end{cases} \label{eq:base}
    \end{align}
    \label{eq:optproblem}
\end{subequations}

This optimization aims to find a robot configuration such that the load born by each joint of the robot does not exceed the upper limit of each joint torque. The optimization is expected to be carried out iteratively between every two sequentially adjacent object poses, and their associated grasp poses. Equation \eqref{eq:rbt_opgoal} is the optimization goal. Equation \eqref{eq:joints} is the constraint on joint ranges. Equation \eqref{eq:ik} is the constraint on the grasping pose, where $FK$ is the forward kinematics. Equation \eqref{eq:joint_tq} is the constraint on joint torques. The torque born by each joint is computed using the robot Jacobian $J$ and the force $F_h$ found by solving the object motion planning problem (equation \eqref{eq:handforce}). Equation \eqref{eq:base} is the constraint on the moving mobile base. We allowed the robot to translate along its local $x$ direction while keeping the other freedoms fixed. There was no particular reason for this constraint. We included it to accelerate optimization. Like object motion planning, the optimization problem may not have a solution. If a feasible robot motion is not found, the grasp pose is determined to be invalid. The system will switch to a different candidate grasping pose and return to the motion planning part to start a new optimization routine.


\section{Experiments and Analysis}
\label{sec:experiments}

\begin{figure}[!htpb]
  \begin{center}
  \includegraphics[width=\linewidth]{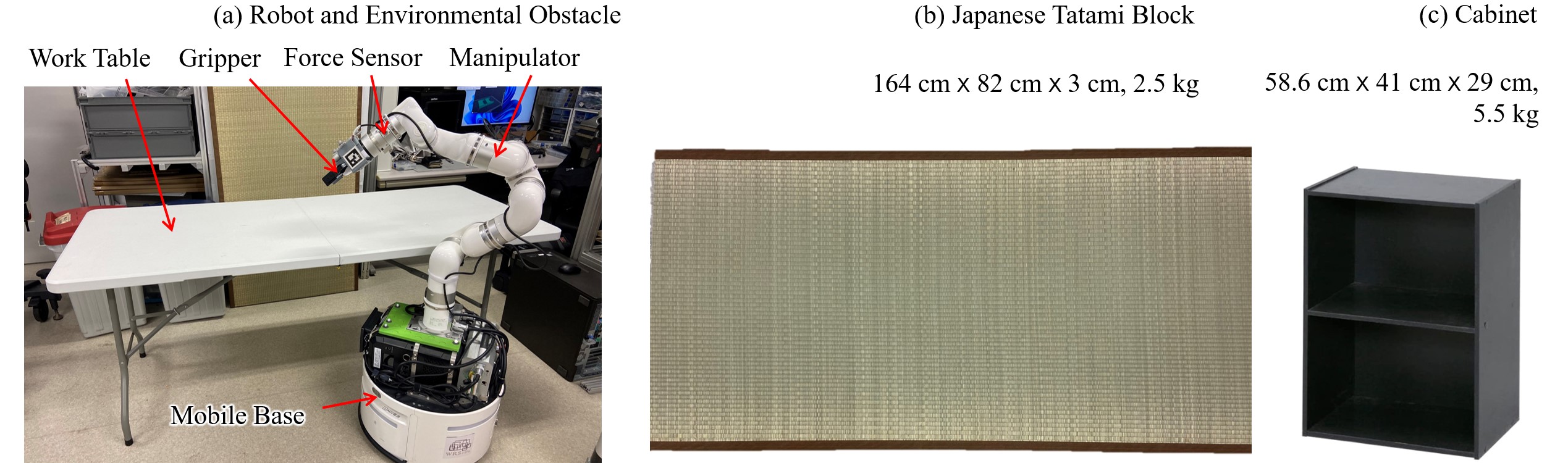}
  \caption{(a) Robot and environmental obstacle (work table). (b) Object 1: Japanese tatami block. (c) Object 2: Cabinet.}
  \label{fig:robots}
  \end{center}
\end{figure}

We carried out experiments and analyses to examine the performance of the method. Notably, we used xArm7 (UFACTORY) for a cooperative manipulator, used Water2 (YUNJI TECHNOLOGY) for a mobile base, and mounted a xArmGripper (UFACTORY) to the xArm7 manipulator as the robot hand. The payload of the xArm7 manipulator is 3.5 kg without any end-effector and 1.87 kg after mounting the hand. The environmental obstacles and objects used in the experiments included a work table (length$\times$width$\times$height$=180$ cm$\times74$ cm$\times75.5$ cm, weight$=12.2$ kg), a Japanese tatami block (length$\times$width$\times$height$=$164 cm$\times$82 cm$\times$3 cm, weight$=2.5$ kg), and a plywood cabinet (length$\times$width$\times$height$=$29 cm$\times$41 cm$\times$29 cm, weight$=5.5$ kg). Fig. \ref{fig:robots} illustrates the robots, environmental obstacles, and objects.

\subsection{Mechanical Compliance}

\begin{figure}[!htpb]
  \begin{center}
  \includegraphics[width=\linewidth]{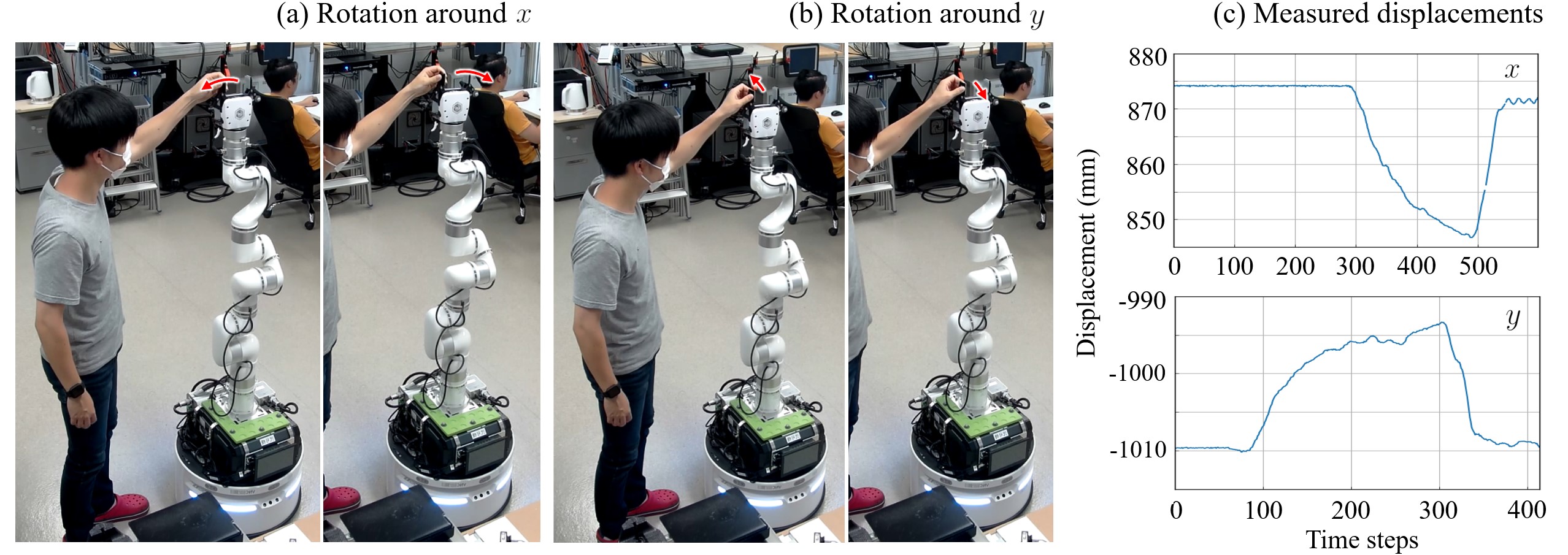}
  \caption{(a) Move the TCP of the vertically posed manipulator around the mobile base's $x$ axis. (b) Move the TCP of the vertically posed manipulator around the mobile base's $y$ axis. (c) Measured displacements. Upper: $x$; Lower: $y$.}
  \label{fig:exp_com}
  \end{center}
\end{figure}

We measured the mechanical compliance of the mobile manipulator following the method presented in Section IV.A. Especially, we set the manipulator part of the robot to a vertical pose, attached optical markers to its TCP, and measured the TCP displacements by recording the changes in marker positions. The setup is shown in Fig. \ref{fig:exp_com}(a) and (b). As illustrated in the figure, we moved the TCP of the vertically posed manipulator around the mobile base's local $x$ and $y$ axes to obtain the ranges of the two assumed orthogonal joints.

Fig. \ref{fig:exp_com}(c) shows the recorded data. Following the data, we concluded that the $\delta d$ in the two directions were 15 mm and 25 mm, respectively. The rotation ranges of the two virtual joints are thus $0.5^\circ$ and $0.859^\circ$ following equation \eqref{eq:dt}.

\subsection{Performance of Object Motion Planning}

\begin{figure}[!htpb]
  \begin{center}
  \includegraphics[width=\linewidth]{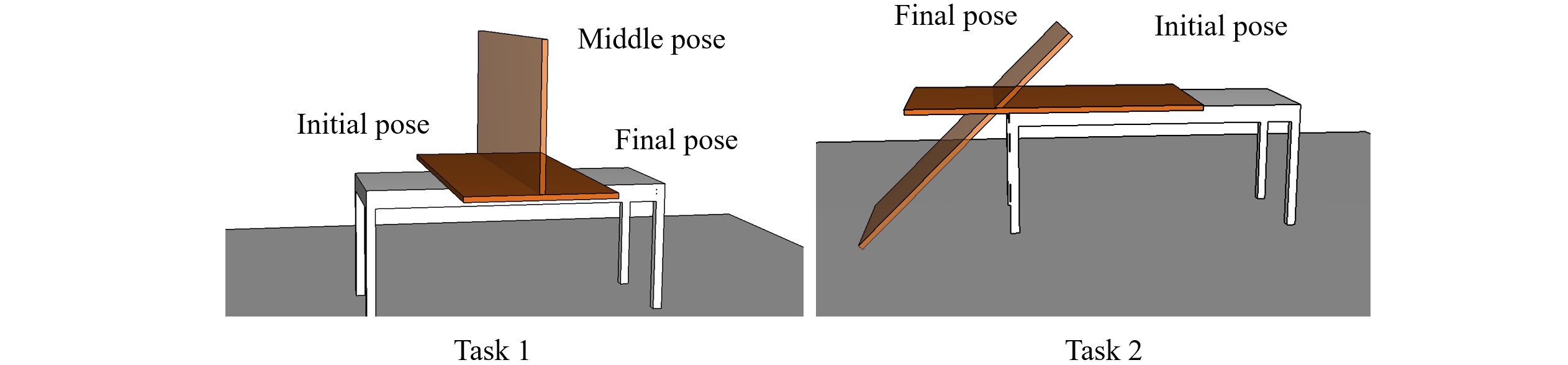}
  \caption{Two tasks used for analyzing object motion planning. Task 1: Flippng the tatami block on the table. Task 2: Unloading the tatami block from the table.}
  \label{fig:tatami_tasks}
  \end{center}
\end{figure}

After obtaining the rotation ranges, we used them to determine the thickness of $\textnormal{M}_\textnormal{e}$ dynamically and thus carried out RRT-based object motion planning. First, we carried out the experiments in simulation using the Japanese Tatami Block and planned the object's motion for the two tasks shown in Fig. \ref{fig:tatami_tasks}. In the first task, the robot planned its motion to flip the tatami on the table. In the second task, the robot planned its motion to unload the tatami block from the table to the ground. Since the tatami is heavier than the robot's maximally affordable payload, the robot must make sure the environment partially supports it during manipulation. The RRT parameters were configured as follows during the experiments: Maximum iterations$=$10000, Smoothing iterations$=$350, and Maximum planning time $=$100 s. 

The left part of Table \ref{table:tasks} shows the results of the first task. To understand the dynamic method's performance, we also plan using fixed $\textnormal{M}_\textnormal{e}$ thickness and compared them with our dynamic method. Columns 1, 2, and 3 of the table are the results with fixed thicknesses. The fixed values were 0.5 cm, 1.5 cm, and 3 cm, respectively. The last column is the results using the dynamically changing thickness computed following the changes of the robot's TCP. The planning time and success rate data are the averages of ten times of attempts. If the distance between the object and the robot exceeded the length of the arm, the thickness was defined as 0. This constrained the trajectory planning to be within the manipulator's operating range. The results show that different fixed thicknesses led to different performances. Thicker ``crust''s will lead to a higher success rate and smaller time costs. When the thickness was 0.5 cm, the success rate was 0.5 cm. Maintaining contact during probabilistic planning under such a thin thickness was very difficult. In contrast, the success rate reaches 90 when the thickness is changed to 3 cm. The success rate of a dynamically changing thickness had a performance competitive to the results of 3 cm thickness. The range of the dynamically changing thickness is from 2.6 cm to 3.35 cm. In the best case, it grew to a value larger than 3 cm. This happened when the robot was most expanded. In the worst case, it reduced to 2.6 cm when the robot was most folded. We could expect the dynamically changing thickness to be more flexible in the best case and safer when the robot is folded.

The right part of Table \ref{table:tasks} shows the results of the second task. The planner successfully solved the problem even though the thickness was fixed to 0.5 cm. However, the success rate was low. Only two of the ten attempts had positive values. Like the first task, the planning time decreased, and the success rates grew as we increased the thickness of $\textnormal{M}_\textnormal{e}$. The proposed dynamic thickness had competitive performance to the 3rd column of the table. The dynamic range differs from the previous task as the robot's TCP positions shifted significantly.

\begin{table}[hbtp]
\caption{Performance of object trajectory planning with different thickness of $\textnormal{M}_\textnormal{e}$}
\label{table:tasks}
\centering
\begin{tabular}{l|cccc|cccc}
    \toprule
    & \multicolumn{4}{c|}{Task 1} & \multicolumn{4}{c}{Task 2}\\
    \cmidrule(lr){2-5} \cmidrule(lr){6-9}
    Thickness (cm) & 0.5  & 1.5  & 3  & 2.60$\sim$3.35 (average 3.00) & 0.5  & 1.5  & 3  & 1.91$\sim$2.38 (average 2.09)\\
    \midrule
    Planning time (s) & -  & 6.34  & 5.82   & 6.29 & 8.39  & 5.81  & 2.40   & 2.92\\
    \midrule
    Success rate (\%) & 0  & 80  & 90   & 90 & 20  & 40  & 60   & 50\\
    \bottomrule
\end{tabular}
\end{table}

\subsection{Real-World Experiments and Torque analysis}

\begin{figure}[!htpb]
  \begin{center}
  \includegraphics[width=\linewidth]{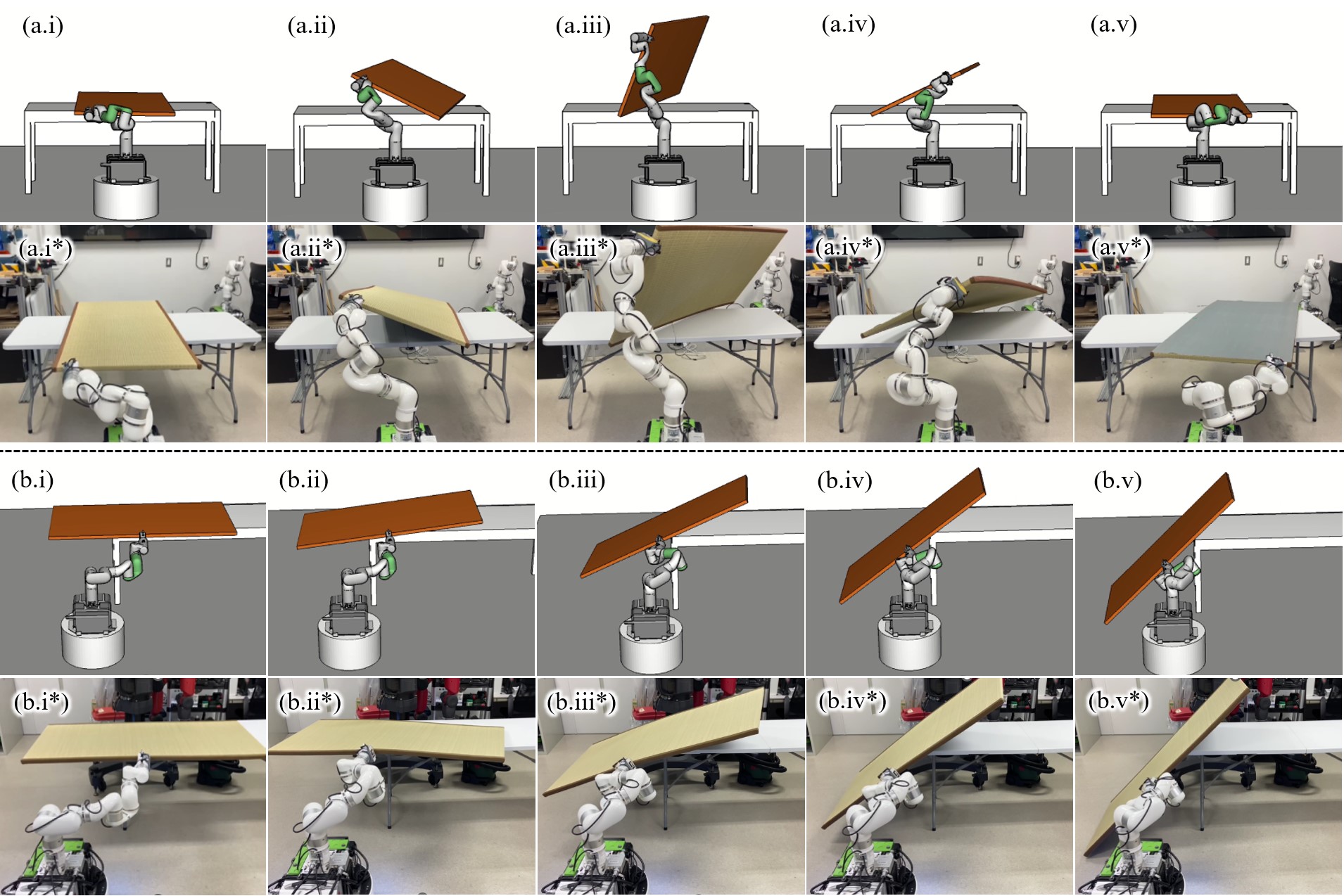}
  \caption{(a) Simulation and real-world results for task 1. (b) Simulation and real-world results for task 2.} 
  \label{fig:flip}
  \end{center}
\end{figure}

Then, we optimized the robot motion for the two tasks based on the object trajectories obtained using the dynamic thickness. The simulation and real-world execution results are shown in Fig. \ref{fig:flip}. The results indicate that the robot maintained contact between the tatami and the environment during the manipulation to let the environment partially share the workload. For task 1, the robot maintained contact between a long edge of the tatami and the table surface. A partial weight of the tatami was supported by the work table, as seen in (a.ii), (a.iii), and (a.iv). For task 2, the robot maintained contact between the tatami surface and the short edge of the table. The table partially supported the tatami until it touched the ground, as seen in (b.iii), (b.iv), and (b.v).

\begin{figure}[!htpb]
  \begin{center}
  \includegraphics[width=\linewidth]{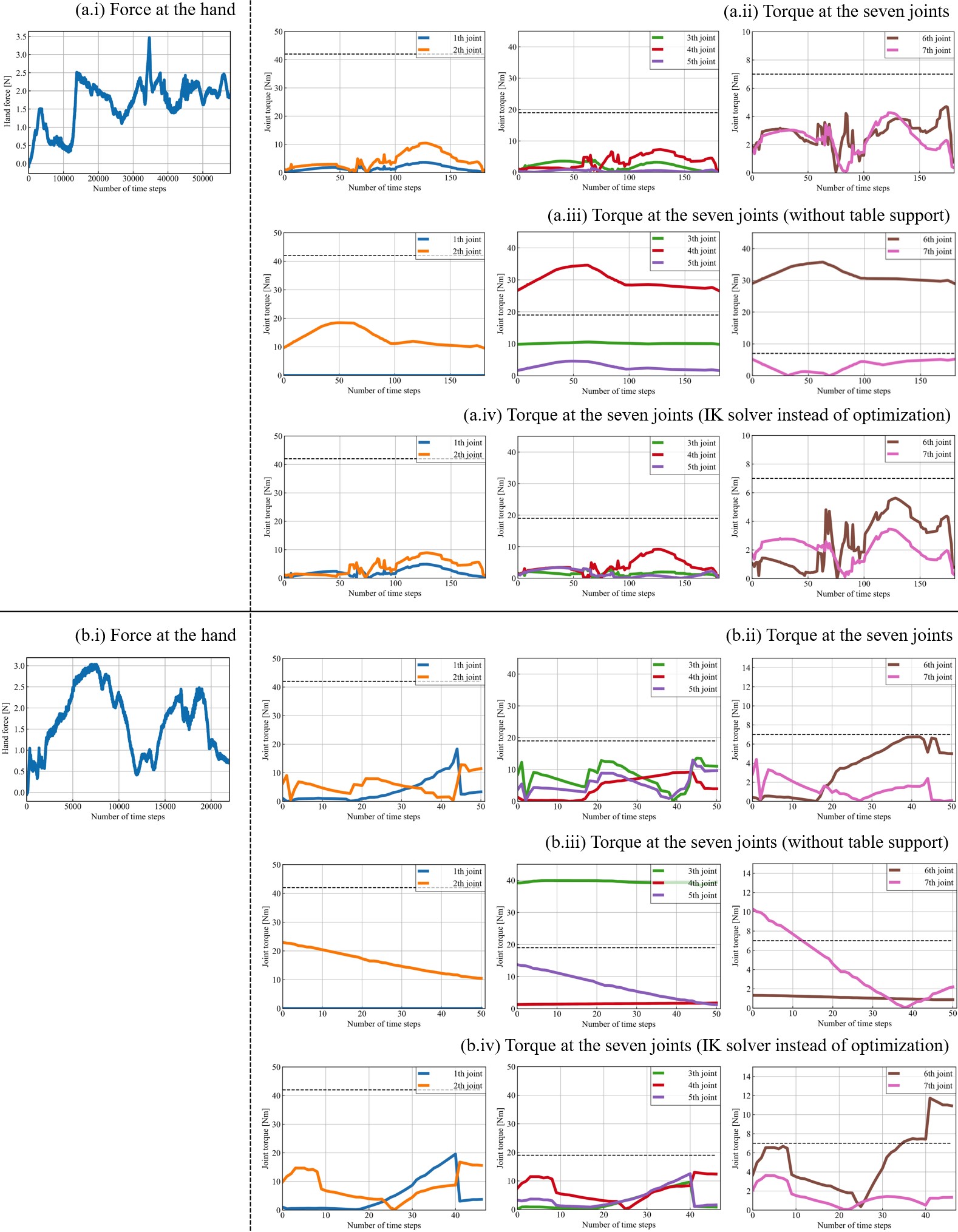}
  \caption{Force and torque born by the hand and joints of the mobile manipulator when performing the two tasks. (a) Task 1. (b) Task 2.} 
  \label{fig:fr}
  \end{center}
\end{figure}

We measured the force born by the robot gripper in the two tasks using the force sensor installed at the end flange of the robot (see Fig. \ref{fig:robots}(a) for the position of the force sensor). The force curves are shown in Fig. \ref{fig:fr}(a.i) and (b.i). The maximum forces born by the robot hand in the two tasks were smaller than 35 N. The robot was bearing a payload smaller than its 3.5 kg payload when performing the tasks. The executions were successful, and the proposed methods are thus considered effective. Fig. \ref{fig:robots}(a.ii) and (b.ii) show the torque at each joint of the robot. The torque curves are separated into three groups and plotted in three different diagrams since joints 1 and 2, joints 3, 4, and 5, and joints 6 and 7 had the same maximum torques, respectively. The dashed lines in the diagrams show the maximum torque that can be born by the joints. The torque curves were coherent with force at hand. All of them were smaller than the maximum torques of the joints.

For comparison, we simulated the torques born by the joints after removing the supporting table. During the simulation, the robot followed the same motion to manipulate the tatami block. Fig. \ref{fig:robots}(a.iii) and (b.iii) show the results. We can observe from the results that without supporting forces from the table, the 5th and 6th joints of the robot in the first task and the 3rd and 7th of the robot in the second task need to provide torque larger than their maximally bearable values. The observation confirmed that environmental support was helpful for the robot to manipulate the tatami block. 

\begin{figure}[!htpb]
  \begin{center}
  \includegraphics[width=\linewidth]{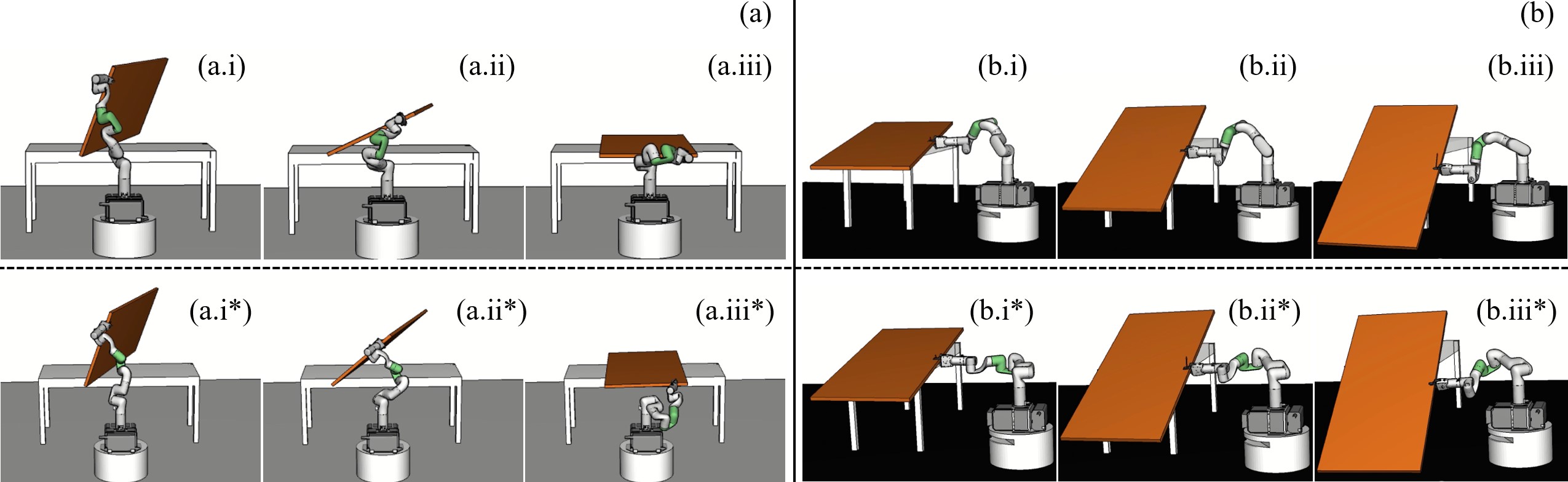}
  \caption{Optimized robot motion (*.i$\sim$iii) $vs.$ motion generated by solving IK sequentially (*.i${}^*\sim$iii${}^*$). (a) Task 1. (b) Task 2. The robots moved following pose sequences that led to smaller joint torque.} 
  \label{fig:copt}
  \end{center}
\end{figure}

We also compared the necessity of robot motion optimization. Fig. \ref{fig:fr}(a.iv) and (b.iv) show the torque curves without solving the optimization problem presented in equation \eqref{eq:optproblem}. The robot motion was generated by directly solving the inverse kinematics of the robot following the trajectory of the object and the holding hand. The curves were slightly different from the ones measured with motion optimization. For task 1, the robot could still successfully finish the execution, although the torques born by the robot joints became more significant than the optimized results. For the second task, the robot failed to carry out the execution. The curves in (b.iv) were simulation results. We could observe from them that the 6th joint of the robot needed to provide a much larger torque to perform the motion successfully. Optimizing the robot motion helped find a proper robot pose sequence and thus avoid excessive joint load. For our readers' convenience, we visualized two optimized and non-optimized robot motion results in Fig. \ref{fig:copt}. Interested readers are recommended to see the figures for detailed examination.

\begin{figure}[!htpb]
  \begin{center}
  \includegraphics[width=\linewidth]{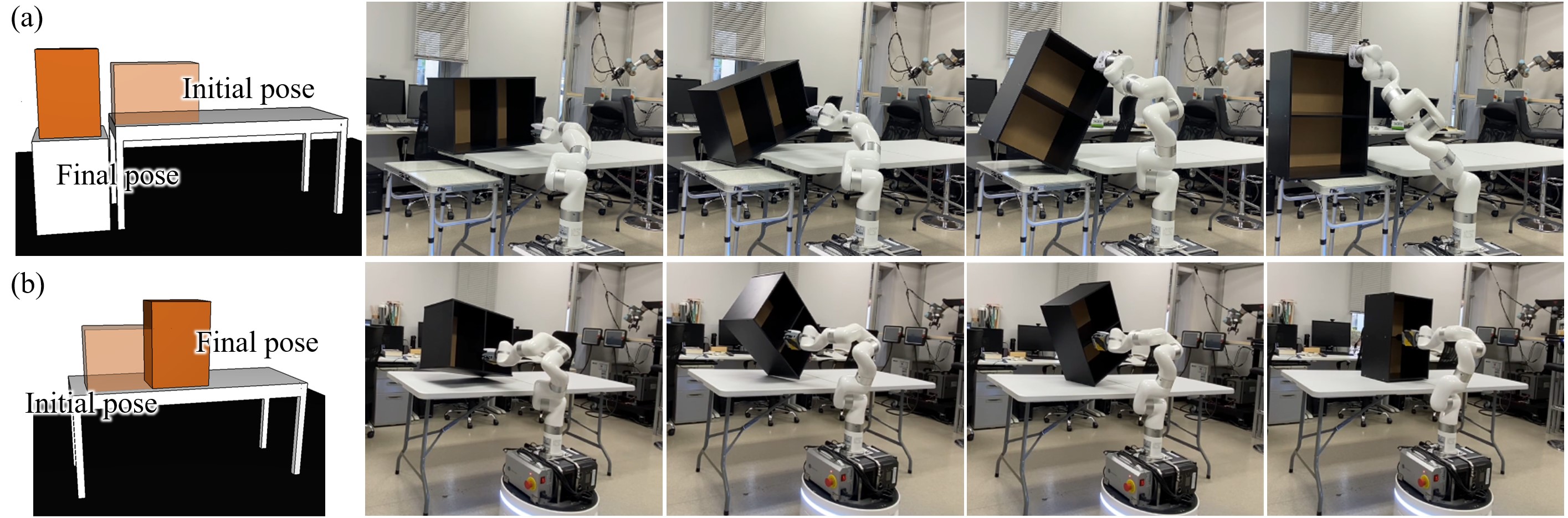}
  \caption{Results with the cabinet. (a) Task 1: Unload the cabinet to a lower table. (b) Task 2: Put the cabinet upright.} 
  \label{fig:cab}
  \end{center}
\end{figure}

In addition to the tatami block, we also conducted experiments using the cabinet shown in Fig. \ref{fig:robots}. Like the tatami block, we defined two tasks shown in Fig. \ref{fig:cab}(a) and (b), and asked the system to find a manipulation sequence for the mobile manipulator to move the cabinet from the initial poses to the goal poses. Since the cabinet is much shorter than the tatami block, we included a smaller and lower table beside the original one and let the robot unload the cabinet from the original table to the lower one. With the help of the proposed method, the robot successfully solved both problems. Readers may also find execution details in the video supplementary published with this manuscript.

\section{Conclusions and Future Work}
\label{sec:conclusions}
We proposed an expanded contact model and used it to plan the trajectory of an object while maintaining contact. The trajectory of the robot based on the trajectory was obtained by solving an optimization problem that incorporates joint torque constraints and other factors. The experimental results show that the method can be applied to various objects and different target states. It helped secure the manipulation of heavy objects by sharing the load between the robot and environmental obstacles. The robot can avoid overloaded joint torque by following the generated manipulation motion. In the future, we plan to verify whether the simulation behavior can be correctly realized in a real environment while improving the problems mentioned in the discussion section.

\bibliographystyle{IEEEtran}
\bibliography{bibfile}

\end{document}